\documentclass[conference]{IEEEtran}

\usepackage[utf8]{inputenc} % upućujete LaTeX da je izvorni kod zapisan u utf8 unicode encodingu
\usepackage{color} % to color the comments

\usepackage{graphicx}
\graphicspath{{figures/}}
\usepackage{subfig}
\usepackage{etoolbox}
\usepackage{mathtools}
\usepackage{amsmath}
\usepackage{graphicx}
\usepackage[table,xcdraw]{xcolor}
\usepackage{array}
\usepackage{overpic}
\usepackage{siunitx}
\usepackage{balance}

\usepackage{tikz}
\usepackage{pgfplots}
\usepackage{color}

\tikzset{every picture/.append style={font=\scriptsize}}
\tikzset{%
  >=latex, % option for nice arrows
  inner sep=0pt,%
  outer sep=2pt,%
  mark coordinate/.style={inner sep=0pt,outer sep=0pt,minimum size=3pt, fill=black,circle}%
}

\usepackage{tikz-3dplot} %requires 3dplot.sty to be in same directory, or in your LaTeX installation
\usetikzlibrary{decorations.markings}
\usetikzlibrary{patterns}
\tdplotsetmaincoords{50}{110}

\begin{document}

\title{Calibration of Heterogeneous Sensor Systems}

\author{\IEEEauthorblockN{Juraj Peršić}
\IEEEauthorblockA{University of Zagreb Faculty of Electrical Engineering and Computing,\\
Department of Control and Computer Engineering,\\
Laboratory for Autonomous Systems and Mobile Robotics (LAMOR)\\
Email: juraj.persic@fer.hr}
}

\maketitle

\begin{abstract}
Environment perception is a key component of any autonomous system and is often based on a heterogeneous set of sensors
and fusion thereof for which sensor sensor calibration plays fundamental role.
It can be divided to intrinsic and extrinsic sensor calibration.
Former seeks for internal parameters of each individual sensor, while latter provides coordinate frame transformation between sensors.
Calibration techniques require correspondence registration in the measurements which is one of the main challenges in the extrinsic calibration of heterogeneous sensors, since generally, each sensor can operate on a different physical principle.
Measurement correspondences can originate from a designated calibration target or from features in the environment. Additionally, environment features can be used to estimate motion of individual sensors and the calibration is found by aligning these estimates.
Motion-based calibration is the most common approach in the online calibration since it is more practical than the target-based methods, although it can lack in accuracy.
Furthermore, online calibration is beneficial for system robustness as it can detect and adjust recalibration of the system in runtime, which can be seen as a prerequisite for long-term autonomy.
\end{abstract}

\IEEEpeerreviewmaketitle

\section{Introduction}

%\im{Općenito mislim da si KDI odlično napisao. Jasno je napisano i dobro oraganizirano. Imam samo par manjih komentara u tekstu koje ćeš vidjeti. Ono što sam najviše radio je spajao paragrafe. Nema potrebe za tolikim dijeljenjem teksta. Ono o čemu bi mogao razmisliti, a čisto je s tehničke strane, je da svaka rečenica započinje u novom redu u \LaTeX sourceu. Na taj način će ti puno lakše biti krosreferencirati pdf sa sourceom. Barem meni, može to ovisiti o editoru.}

Robust environment perception is one of the essential tasks which an autonomous mobile robot or vehicle
has to accomplish. To achieve this goal, various sensors such as cameras, radars, LiDARs, and inertial navigation units are used and information thereof is often fused. Essential tasks such as simultaneous localization and mapping (SLAM), detection and tracking of moving objects, odometry, etc. are often improved by sensor fusion.

A fundamental step in the fusion process is sensor calibration, both intrinsic and extrinsic.
Former provides internal parameters of each sensor (e.g. focal length of a camera, bias in LiDAR range measurements), while latter provides relative transformation from one sensor coordinate frame to the other. The calibration can tackle both parameter groups at the same time or assume that sensors are already intrinsically calibrated and proceed with the extrinsic calibration. Additionally, temporal synchronization of the sensors is sometimes performed within the calibration.

Intrinsic parameters are related to the working principle of the sensor. Therefore, methods for finding intrinsic parameters do not share much similarities for different types of sensors.
On the other hand, parametrization of extrinsic calibration, i.e. homogeneous transformation, can always be expressed in the same manner, regardless of the sensors involved in it. Despite that, solving the extrinsic calibration requires finding correspondences in the data acquired by the sensors which can be challenging since different types of sensors measure different physical quantities.

After correspondence registration, optimization steps are performed to estimate the calibration parameters. While some methods require intrinsically calibrated sensors to find the extrinsic calibration, others perform optimization on both parameter groups simultaneously. These methods typically try to satisfy some geometric constraints through minimization of a problem-specific reprojection error. The geometric constraints involve nonlinearities which often cannot be solved analytically. To resolve that problem, estimators use iterative techniques to find the appropriate solution. Due to the nonconvexity of the problem caused by the nonlinearities, these methods have a risk of converging to a local minimum. To avoid that risk, some methods divide optimization in initial rough estimates that guarantee near-optimal solutions followed by nonlinear iterative refinement step. 
The success of the optimization is highly dependant on the provided data. Important step before the data acquisition is to determine minimal requirements on the dataset for which the problem becomes identifiable (or observable in case of dynamical systems).% \im{Nema potrebe za ovako malim paragrafima.}

The calibration approaches can be target-based or targetless.
In the case of target-based calibration, correspondences originate from a specially designed target, while
targetless methods utilize environment features perceived by both sensors.
Former has the advantage of the freedom of design which maximizes the chance of both sensors perceiving the calibration
target, but requires the development of such a target and execution of an appropriate offline calibration procedure.
The latter has the advantage of using the environment itself as the calibration target and can operate online by
registering structural correspondences in the environment, but requires both sensors to be able to extract the same
environment features.
Registration of structural correspondences can be avoided by motion-based methods, which use the system's
motion estimated by the individual sensors to calibrate them. These methods have two main advantages: (i) they rely less on the sensors operating principles and can be applied to different sensors, assuming that a sensor can estimate its motion, (ii) unlike other methods, they are able to extrinsically calibrate sensors with non-overlapping fields of view.

%However, for all practical means and purposes, the targetless methods are hardly feasible due to limited resolution of current automotive radar systems, as the radar is virtually unable to infer the structure of the detected object and
%extract features such as lines or corners.
%Novo profesorovo
%Therefore, we focus our research on target-based methods.
%Staro Ivanovo
%Since an automotive radar is used in the present paper and we are targeting in situ calibration
%techniques, our approach will focus further on target-based methods.

%Calibration targets are designed to suit the working principle of the sensor.

The paper is organized as follows.
Section~\ref{tbm} elaborates the calibration methods including calibration targets. Section~\ref{tlm} explains common approaches in targetless calibration which uses features in the environment as correspondences in the sensor data.
Afterwords, Section~\ref{mbm} provides details on existing motion-based methods. Lastly, conclusion is given in Section~\ref{con} with focus on the importance of sensor calibration.
%\jp{The paper is organized as follows.

\section{Target-based methods}\label{tbm}

Calibration targets are frequently used due to numerous advantages. They simplify the correspondence registration step since the number and type of correspondences is known in advance which virtually eliminates the problems associated with outliers. Additionally, target-based methods can use a priori knowledge about the target which can enhance the calibration results. Therefore, target-based methods are generally more precise than the targetless. Finally, there are no requirements on the environment which can be uninformative and prevent targetless methods from success. However, it is the least practical method since it requires design and construction of the target and may not always be suitable (e.g. end-user applications like smartphones). Furthermore, it has to be performed offline before any other application for which the calibration is important. Therefore, it is impossible to make any runtime adjustments and the process has to be repeated in case of decalibration.

Properties of a well-designed target are (i) ease of detection and (ii) high localization accuracy for all the sensors in the calibration. The former ensures the success of the correspondence registration, while the latter has great influence on the quality of the results given by the optimization step. Furthermore, if the a priori knowledge about the target is used, construction imprecision may lead to poor calibration results.
Perception sensors used in robotics utilize a wide range of physical phenomena to extract information about the environment. Due to different type of data provided by heterogeneous sensors, there exist many diverse target designs. In the sequel we will present some of the designs grouped by the sensor types.

\subsection{Camera}

Cameras are passive sensors that utilize the light which goes through the lens and is detected at the optical sensor. They are a rich source of information with an affordable price what makes them commonly used in robotics and other fields.
Due to their long presence and frequent usage, intrinsic camera calibration has been given a lot of research attention which resulted in many camera description models and calibration techniques. While cameras with high distortion such as fisheye and omnidirectional cameras require more complex models \cite{Scaramuzza2008}, commonly used cameras with slight distortion are usually modelled as pinhole cameras with a previously rectified image. This intrinsic parametrization consists of distortion coefficients (e.g. radial distortion) and camera matrix formed by focal length, pixel scale factors, principal point and skewness between the axis.

Commonly used camera calibration targets are planar checkerboard patterns. They are suitable because they can be easily detected in the image and enable sub-pixel resolution using interpolation based on a known target dimensions. Calibration methods are based on pioneering work by Tsai \cite{Tsai1987} and Zhang \cite{Zhang2002}. Besides checkerboard pattern, a grid of circles is also frequently used \cite{Heikkila1997} with comparison study of different patterns given in \cite{Mallon2007}.
Novel calibration target is presented in \cite{Li2013} where authors use a noise-like pattern with many features of varying scales. 
It is suitable for both intrinsic and extrinsic calibration of multiple cameras with no or little filed of view (FOV) overlap.
The only requirement is that the neighbouring cameras observe parts of the target which may not overlap at all. 
Additionally, it can simultaneously handle both close-range and far-range cameras.

\subsection{LiDAR}

LiDARs are active sensors that use light pulses to determine the range of the objects in the environment and provide results in the form of point cloud, i.e., a set of 3D points. They usually consist of one or more rotating light transmitters/receivers. LiDARs are classified as: 1D when they measure a distance on a single ray, 2D when they measure distances in only one plane, and 3D or multi-layer when they measure distances in multiple planes. Considering sensor calibration, 2D and 3D LiDARs have received extensive attention due to the application requirements and possibility to recover structure from the environment.

Intrinsic parameters of interest are range measurement offsets and pose of the individual rays to the common reference frame. Unlike cameras, precision of intrinsic factory calibration parameters is usually considered sufficient. However, for the applications that require higher precision, authors in \cite{Jimenez2011} propose a method for intrinsic calibration of rotating 3D LiDAR using a box with known dimensions, while the authors in \cite{Muhammad2010} use a planar wall as a calibration target.

Extrinsic calibration between multiple LiDARs is mostly done by motion-based methods which are applicable when there is small or none FOV overlap. For sensor configurations in which mutliple 2D LiDARs share the same parts of FOV, Fernandez-Moral et al. \cite{Fernandez-Moral2015} presented a solution which uses corner structures to perform extrinsic calibration. Additionally, using the rank of Fisher Information Matrix (FIM) they show that problem becomes identifiable when at least three perpendicular planes are observed.

\subsection{LiDAR\,--\,Camera}

Point clouds from 2D/3D LiDARs are often fused with camera images. Both are rich information sources and precise extrinsic calibration is crucial for tasks such as 3D reconstruction what led to development of many calibration methods.
A common approach in target-based LiDAR\,--\,camera calibration is using planar targets which are easily detected and localized in the point cloud covered by a pattern (e.g. checkerboard) which allows estimation of the plane position and orientation in the image.

Widely adopted and extended method presented by Zhang and Pless \cite{Zhang2004} introduced point-plane geometric constraint initially designed for 2D LiDAR\,--\,camera calibration. LiDAR points originating from the target plane are transformed into the camera frame and the method tries to minimize point to plane distances based on the estimated plane parameters in the image. Pandey et al. \cite{Pandey2010} showed that the method is also applicable in case of 3D LiDAR\,--\,camera calibration. Zhou and Deng \cite{Zhou2012} improved the method by introduction of additional constraints which decoupled rotation from translation.
They achieved better results than other methods because their method is less affected by errors in the plane parameters estimation in the checkerboard image. Additionally, they showed that for a 2D LiDAR, at least five correspondences should be made with different target orientations, while a 3D LiDAR required minimum of 3 different views. Geiger et al. \cite{Geiger2012} tried to reduce time of the calibration procedure by extending the method with global correspondence registration which allows for multiple plane observations in a single shot.
The same constraint was used by Mirzaei et al. in \cite{Mirzaei2012} where instead of checkerboard patter, AprilTag fiducial markers were used \cite{Olson2011}. Additionally, they extended the extrinsic calibration with estimation of intrinsic LiDAR parameters. AprilTag markers and the same geometric constraint were also used in \cite{Owens2015} as a part of multi-sensor graph based calibration.

%For example, extensive research exists on 3D LiDAR-camera calibration with a planar surface covered by a
%chequerboard \cite{Pandey2010,Geiger2012,Zhou2012} or a set of QR codes \cite{Mirzaei2012,Owens2015}.
%Extrinsic calibration of a 2D LiDAR-camera pair was also calibrated with the same target \cite{Zhang2004}, while
%improvements were made by extracting centerline and edge features of a V-shaped planar target \cite{Kwak2011}.

Besides commonly used point-plane constraint, 3D LiDAR-camera pair was calibrated based on the point-point correspondences. Velas et al. \cite{Velas2014} used a target with circular holes which allows a single-shot calibration and does not require observation of the plane in multiple orientations. Similar geometric constraints where used by Kwak et al. \cite{Kwak2011} for 2D LiDAR-camera calibration. Improvements were made by extracting centreline and edge features of a V-shaped planar target.
Furthermore, an interesting target adaptation to the working principle of different sensors was presented by Bormann et al. \cite{Borrmann2012}, where the authors proposed a method for extrinsic calibration of a 3D LiDAR and a thermal camera by expanding a planar checkerboard surface with a grid consisting of light bulbs.

\subsection{Radar\,--\,Camera/Lidar}

Radars are active sensors which, similarly to the LiDAR, emit an electromagnetic signal and determine the range of objects in the vicinity based on the returned echo. Although being frequently used in automotive applications due to their low price and robustness, extrinsic radar calibration has not gained much research attention. The existing methods are all target-based since, for all practical means and purposes, the targetless methods are hardly feasible due to limited resolution of current automotive radar systems, as the radar is virtually unable to infer the structure of the detected object and extract features such as lines or corners.
Current radars have no elevation resolution while the information about the detected objects they provide contains range, azimuth angle, radar cross section (RCS) and range-rate based on the Doppler effect. Although having no elevation resolution, radars have substantial elevation FOV which makes the extrinsic calibration challenging due to the uncertainty of the measurements.

Concerning automotive radars, common operating frequencies (24\,GHz and 77\,GHz) result with reliable detections of conductive objects, such as plates, cylinders and corner reflectors, which are then used in intrinsic and extrinsic
calibration methods \cite{Knott1993}.
Wang et al. \cite{Wang2011} used a metal panel as the target for radar\,--\,camera calibration.
They assume that all radar measurements originate from a single ground plane, thereby neglecting the 3D nature of the
problem.
The calibration is found by optimizing homography transformation between the ground and image plane.
Contrary to \cite{Wang2011}, Sugimoto et al. \cite{Sugimoto2004} take into account the 3D nature of the problem.
% They manually search the field of view (FoV) of the radar with a corner reflector to find intensity maximums.
Therein, they manually search for detection intensity maximums by moving a corner reflector within the FOV. % of the radar.
They assume that detections lie on the radar plane (zero elevation plane in the radar coordinate frame).
Using these points a homography transformation is optimized between the radar and the camera.
The drawback of this method is that the maximum intensity search is prone to errors, since the
return intensity depends on a number of factors, e.g., target orientation and radar antenna radiation pattern which
is usually designed to be as constant as possible in the FOV.

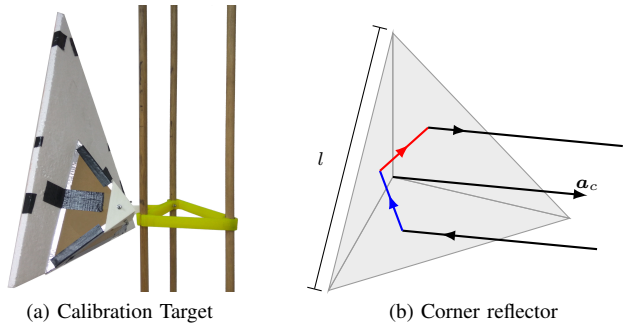
\begin{figure}[!t]
   \subfloat[Calibration Target]{
   \includegraphics[height=3.8cm,keepaspectratio]{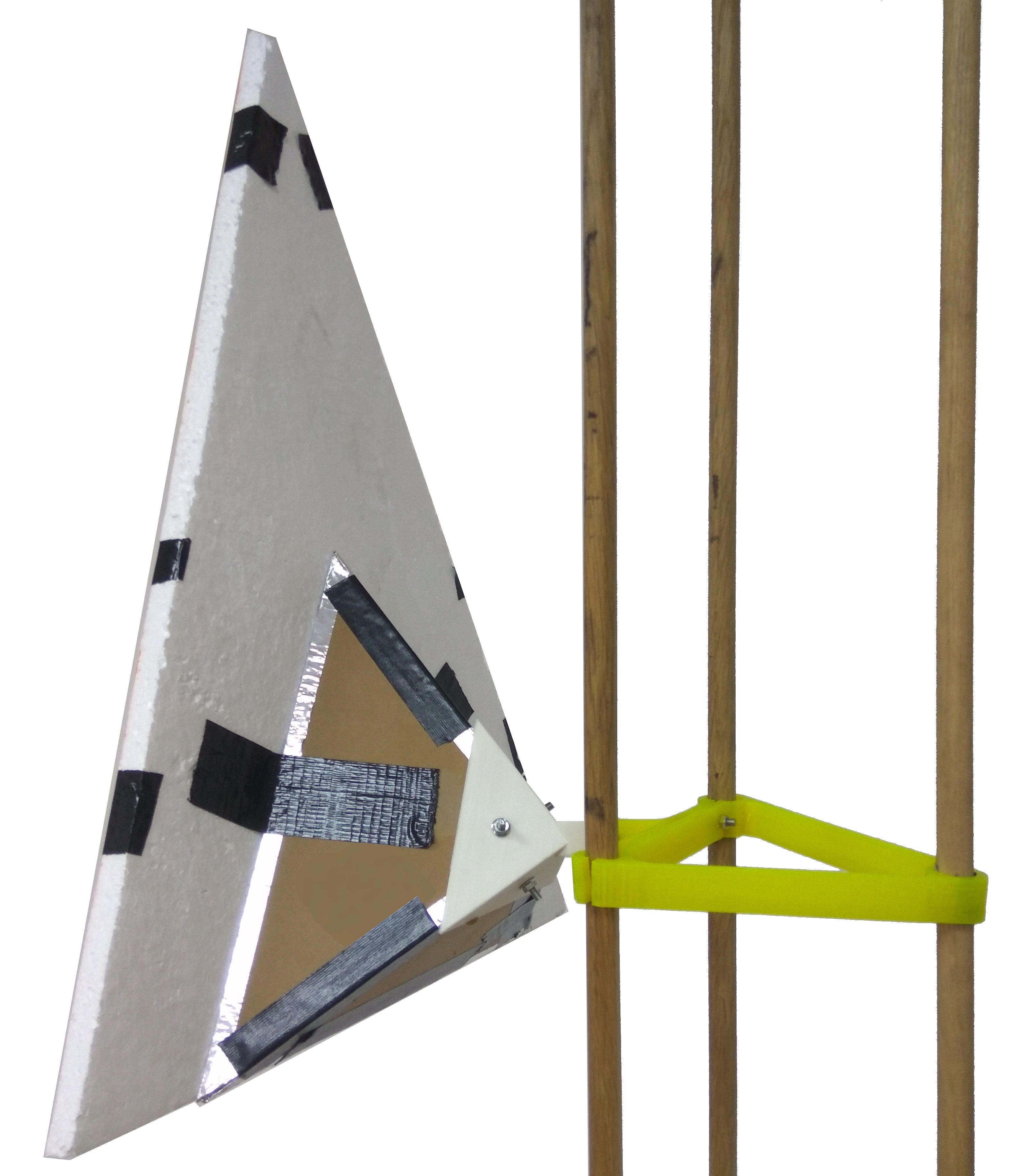}
\label{fig:calibration_target}}
   \qquad
   \subfloat[Corner reflector]{
   \begin{tikzpicture}[scale=2.5,tdplot_main_coords]
\tdplotsetmaincoords{50}{110}
%\tdplotsetmaincoords{70}{0}
%set up some coordinates 
%-----------------------
\coordinate (O) at (0,0,0);

%\draw[thick,->] (0,0,0) -- (1.5,0,0) node[anchor=north east]{$x$};
%\draw[thick,->] (0,0,0) -- (0,1.5,0) node[anchor=north west]{$y$};
%\draw[thick,->] (0,0,0) -- (0,0,1.5) node[anchor=south]{$z$};

%determine a coordinate (P) using (r,\theta,\phi) coordinates.  This command
%also determines (Pxy), (Pxz), and (Pyz): the xy-, xz-, and yz-projections
%of the point (P).
%syntax: \tdplotsetcoord{Coordinate name without parentheses}{r}{\theta}{\phi}

%draw figure contents
%--------------------

%draw the main coordinate system axes
\pgfmathsetmacro{\sideLength}{1}
\tikzstyle{triangleFill} = [fill=gray!29,fill opacity=0.5, draw=gray!79]
\filldraw[triangleFill](0,0,0)--(\sideLength,0,0)--(0,\sideLength,0)--cycle;
\filldraw[triangleFill](0,0,0)--(0,0,\sideLength)--(0,\sideLength,0)--cycle;
\filldraw[triangleFill](0,0,0)--(0,0,\sideLength)--(\sideLength,0,0)--cycle;

%\filldraw[fill=gray!39,fill opacity=0.5](0,\sideLength,0)--(0,0,\sideLength)--(\sideLength,0,0)--cycle;
%\draw[thick] (0,0,0) -- (1,0,0);
%\draw[thick] (0,0,0) -- (0,1,0);
%\draw[thick] (0,0,0) -- (0,0,1);
%\draw[thick] (1,0,0) -- (0,0,1);
%\draw[thick] (0,1,0) -- (0,0,1);

\tdplotsetcoord{Pca}{3}{54.7}{45}
\draw[thick,->] (0,0,0) -- (Pca) node[anchor=south]{$\boldsymbol{a}_c$};

\coordinate (P1) at (2.1313,1.9313,1.7336);
\coordinate (P2) at (0.4,0.2,0);
\coordinate (P3) at (0.2,0,0.20026);
\coordinate (P4) at (0,0.2,0.40053);
\coordinate (P5) at (1.7313,1.9313,2.1341);

\tikzset{->-/.style={decoration={
  markings,
  mark=at position #1 with {\arrow{>}}},postaction={decorate}}}
  
\draw[thick,->-=.8] (P1) -- (P2);
\draw[thick,color=blue,->-=.6] (P2) -- (P3);
\draw[thick,color=red,->-=.6] (P3) -- (P4);
\draw[thick,->-=.2] (P4) -- (P5);

%\draw[dashed, color=blue] (O) -- (rFOVup);
%\draw[dashed, color=blue] (O) -- (rFOVdown);

%{$\theta$}

%draw some dashed arcs, demonstrating direct arc drawing
%\draw[dashed,tdplot_rotated_coords] (\rvec,0,0) arc (90:135:\rvec);
%\draw[dashed] (\rvec,0,0) arc (0:45:\rvec);

%set the rotated coordinate definition within display using a translation
%coordinate and Euler angles in the "z(\alpha)y(\beta)z(\gamma)" euler rotation convention
%syntax: \tdplotsetrotatedcoords{\alpha}{\beta}{\gamma}

\pgfmathsetmacro{\offset}{0.2}
\pgfmathsetmacro{\offsetX}{0.02}
\draw[|-|] (\offset,0,\sideLength+\offset)--(\sideLength+\offset+\offsetX,0,\offset-\offsetX); node[right]{$$};
\node[tdplot_screen_coords,text width=1cm] at (-0.2,0.1) {$l$};

\end{tikzpicture}
   \label{fig:cornerReflector}}
   \caption{Constructed calibration target from \cite{Persic2017} and the illustration of the working principle of the triangular trihedral corner reflector}
   \vspace*{-0.5cm}
\end{figure}

%In \cite{Stanislas2015} radar performance is evaluated using a 2D LiDAR as a ground truth with a target composed of radar tube
%reflector and a square cardboard.
%The cardboard is practically invisible to the radar, while enabling better detection and localization in the LiDAR point
%cloud.
%These complementary properties were taken as an inspiration for our target design.
While the above described radar calibration methods provide sufficiently good results for the targeted applications,
they lack the possibility to fully assess the placement of the radar with respect to other sensors.
Research on 3D LiDAR-radar calibration was conducted by Peršić et al. \cite{Persic2017}. They propose a method which estimates a 6 degress of freedom (DOF) extrinsic calibration of a 3D LiDAR\,--\,radar pair. The method includes a target design suitable both for the LiDAR and the radar shown in Fig.~\ref{fig:calibration_target}. It is inspired by a target constructed by Stanislas and Peynot \cite{Stanislas2015} where radar performance is evaluated using a 2D LiDAR as a ground truth with a target composed of radar tube reflector and a square cardboard. Target for 3D LiDAR\,--\,radar calibration consists of a styrofoam triangle which is invisible to the radar while it has good properties for detection and localization in the LiDAR point cloud. Radar receives the echo from the trihedral corner reflector shown in Fig.~\ref{fig:cornerReflector} which has high RCS and low orientation sensitivity.
In the end, extrinsic calibration parameters are found by two-step optimization. The first step is based on the reprojection error minimization while the second uses space distribution of RCS, measure of the detection intensity, to estimate variables which are not observable from the reprojection error due to the lack of radar's vertical resolution.

\section{Targetless methods}\label{tlm}

In order to maintain reliability of a perception system, sensor calibration has to be performed occasionally. Sensors displacement due to mechanical vibrations, changes of intrinsic parameters due to the variable environment conditions such as temperature and pressure, are some of the effects that can cause sensor decalibration. In such cases target-based methods are impractical and can restrict usability of the system which led to development of the targetless methods. They eliminate the need for artificial targets by using environment features to match correspondences in the sensor data.
This problem is especially challenging in the heterogeneous sensor systems. It is feasible when sensors provide enough information about the environment to extract its structure. Therefore, this techniques, described in the sequel, are mainly used in camera and LiDAR calibration.

\subsection{Camera}

Barazzetti et al. \cite{Barazzetti2011} proposed an approach for intrinsic camera calibration using only natural scenes. Their method uses feature extraction methods and robust estimation techniques to create correspondences between different views of the same scene. 
It is suitable for scenes with many features that can be uniquely described.
However, repetitive textures (e.g. building facades, tiles) result in image features with similar descriptors which can be easily mismatched and thus 
%can cause incorrect matching  of the image features since they have similar descriptors which 
compromise the calibration results.
Although showing valuable results, authors conclude that high precision and industry applications still require target-based methods for desired calibration accuracy. Similar approach was adopted by Fraser and Stamatopoulos \cite{Fraser2014} where they showed comparable results to the target-based methods.

In order to retrieve depth information about the environment, two cameras are often rigidly connected to form a stereo vision system. Besides the intrinsic calibration of individual cameras, high precision of extrinsic calibration between the cameras is crucial for successful stereo reconstruction. Common methods for intrinsic target-based can be used to obtain the extrinsic calibration. Online targetless calibration is a greater challenge \cite{Hansen2012,Spangenberg2013,Thacker2005} and Ling and Shen \cite{Ling2016} have presented an online targetless approach. It minimizes epipolar errors between the image pairs based on the sparse natural features to obtain 5 degrees-of-freedom (DOF) transformation between the cameras. Similarly as in monocular vision, scale of the translation vector is unobservable. They show comparable results to the target-based methods.

\subsection{Camera\,--\,LiDAR }
%\im{Kod odnosa treba biti ili en-dash ili em-dash. Obična crta se koristi u primjerima kao što vidiš u prethodnoj rečenici. :-)}
Informativeness of these sensors enables inference on the structure of the environment that can be used in generating correspondences. For example, Levinson and Thrun \cite{Levinson2013} based their calibration of a 3D\,--\,LiDAR and a camera on line features detected as intensity edges in the
image and depth discontinuities in the point cloud. Their method is able to detect decalibration on-the-fly and track the gradual drift of the sensor pose over time. Similar approach was adopted by Moghdam et al. \cite{Moghadam2013} where they increased the robustness of their method by handling one-to-many correspondence registration by re-weighting the error metric. Gong et al. \cite{Gong2013} proposed an approch in which they use arbitrary trihedrons commonly found in urban and indoor environments (e.g. corners of the buildings).

In addition to range measurements, LiDARs also provide information about returned signal's intensity. Pandey et al. \cite{Pandey2015} find extrinsic calibration by maximizing the mutual information between the cameras grayscale pixel intensities and projected surface reflectivity values measured by the LiDAR. For success of their method it is important to first perform intrinsic inter-beam calibration of the surface reflectivity values. The concept of mutual information was also used by Taylor and Nieto \cite{Taylor2013}. Instead of using dense information from the point cloud, they only project selected features into the 2D LiDAR image. Additionally, they complement returned intensity information with estimated surface normal as there exist strong statistical dependence between these quantities. They show that the method is applicable to variety of LiDARs. Furthermore, mutual information between camera image and LiDAR generated reflectance image was used by Napier et al. \cite{Napier2013} to calibrate a push-broom 2D LiDAR with camera in natural scenes. The method allows calibration of sensors without overlapping FOVs, but it requires ego-motion information.

Generation of 2D image from the LiDAR's point cloud was also done by Scaramuzza et al. \cite{Scaramuzza2007a}. Instead of intensity, they introduced bearing angle images which are constructed from angle difference of the surface normals in the point cloud. This metric highlights environment plane intersections arising from wall corners and other similar discontinuities. However, their method requires manual registration of the correspondences.
Lastly, extracting features from the environment can lead to a high number of correspondences. Scott et el. \cite{Scott2016} claimed that not all correspondences are equally informative and that appropriate choice of scenes can improve calibration. They use normalised information distance as a criteria for scene selection scheme which provided more effective and precise calibration results using a few scenes.

\section{Motion-based methods}\label{mbm}

Motion-based calibration techniques compare ego-motion estimates from individual sensors to perform calibration. These methods can be classified as targetless methods because they also use environment features, but since they are only used to estimate ego-motion, developed methods are applicable to a wider range of sensor configurations. Only requirement is that the sensor can estimate its motion. Moreover, for the sensors such as IMU or encoder odometry, motion-based methods are the only viable solution. Additionally, they are virtually the only option for calibration of sensors whose FOVs do not overlap. %\im{Ovo je vrlo vjerojatno točno u većini slučajeva, ali zar nije ona jedna metoda s ETH-a koja može kalibrirati kamere čiji se FOV ne presjeca?} 
Many of these methods are agnostic in terms of the sensor choice. In the sequel, some of the general methods will be addressed, followed by methods which have some special contribution in specific sensor configurations.

\subsection{General Methods}

In their work \cite{Schneider2013}, Schneider et al. have proposed a solution for extrinsic calibration of sensors based on Unscented Kalman Filter. The method is generic and can be used with sensors which provide both 3DOF and 6DOF motion estimates. However, they require time-synchronized delta poses.
Furgale et al. \cite{Furgale2013,Rehder2016} have proposed a method which relaxes that constraint by using continuous-time batch estimation while simultaneously estimating both spatial and temporal calibration parameters.
Brookshire and Teller \cite{Brookshire2012} proposed an approach in which they explicitly model the noise via the Lie algebra yielding a constrained FIM from which they analyse motion degeneracy and proceed to singularity-free optimization procedure.
FIM was also used by Maye et al. \cite{Maye2015} to detect unobservable directions in parameter space from the available data. Additionally, they used information gain as a measure upon which they accept new measurements into the batch optimization. By reducing the total number of correspondences they created a framework for feasible online calibration.

Furthermore, Huang and Stachniss \cite{Huang2017} have addressed the problem of high measurement noise which compromises the results of the commonly used least square optimization techniques. They improved the calibration results by adopting Gauss-Helmert optimization paradigm which jointly optimizes calibration parameters and pose observation errors.

\subsection{Camera - IMU}

Visual-inertial odometry is able to accurately estimate the 6DOF motion and it is well suited for many robotic tasks. However, it requires precise extrinsic calibration. Mirzaei and Roumeliotis \cite{Mirzaei2008} proposed a Kalman filter based approach for IMU\,--\,camera calibration. They based their method on estimating the camera motion using checkerboard pattern. Through the observability analysis based on the Lie derivatives rank criterion they showed that it is necessary to excite at least two rotational axis of the system to make the calibration parameters observable. Kelly and Sukhatme \cite{Kelly2011} have continued on the previous research by discarding the checkerboard and using environment features for the visual odometry. They showed that additional two translational axes need to be excited in order to resolve camera scale issue and make the calibration parameters observable.
Keivan and Sibley \cite{Keivan2015} proposed a SLAM solution which is able to detect system decalibration and perform calibration online. Compared to the previously described methods, they are additionally able to recalibrate camera's intrinsic parameters. They evaluated the method on an experiment where they doubled the focal length on-the-fly.

\subsection{Hand\,--\,Eye calibration}

Many robotic applications involve manipulator equipped with a wrist-mounted sensor such as camera. Calibration between the end of the manipulator and the perceptive sensor is crucial in these applications. This problem is often referred to as an $AX=XB$ problem due to the emerging equation that needs to be solved ($A$ and $B$ represent manipulator and sensor movement, respectively, while $X$ represents the extrinsic calibration). It has been studied for more than three decades \cite{Shiu1987a} and many solutions exist.

Some of the recent advances in the field have dealt with the problem of unknown correspondences caused by asynchronous sensors or missing detections \cite{Ackerman2013a, Ma2016a}. 
Furthermore, a general solution for motion-based extrinsic and temporal calibration was given by Taylor and Nieto \cite{Taylor2016}. The solution is based upon the framework of $AX=XA$ problem which is further enhanced by targetless methods if the sensor types and overlaps allow such refinements. It was evaluated through calibration of several vehicle-mounted sensor configurations.

\section{Conclusion}\label{con}

Robotic systems are increasingly adopting heterogeneous multi-sensor approaches for which correct sensors calibration is a necessity, both intrinsic and extrinsic. Nowadays, robotic applications use a wide spectrum of sensors for which many calibration methods have been developed. While some solve specific sensor configurations, others aim at more general solutions. In this overview, some of the representative methods were presented for commonly used sensors such as cameras, LiDARs, IMUs and radars. Offline calibration has been studied for a long time, while the field of online calibration has been given a special focus only as of lately. Online detection and correction of system decalibration is crucial for robustness of any autonomous system which makes it one of the prerequisites for reliable long-term autonomy.

%\im{Ostao ti je u bibu na jednom mjestu link na članak.}

%\section*{Acknowledgment}

%\newpage
%\newpage
\balance
\bibliographystyle{IEEEtran}
\bibliography{library}
\end{document}